\def\year{2021}\relax
\documentclass[letterpaper]{article} 
\usepackage{aaai21}  
\usepackage{times}  
\usepackage{helvet} 
\usepackage{courier}  
\usepackage[hyphens]{url}  
\usepackage{graphicx} 
\usepackage{natbib}  
\usepackage{caption} 
\usepackage{amsmath}
\usepackage{amssymb}
\usepackage{amsfonts}
\usepackage{enumitem}
\usepackage{graphicx}
\usepackage{xcolor}
\graphicspath{{Figure/}}
\title{A Unified Pre-training Framework for Conversational AI}
\author{Siqi Bao\thanks{~Equal contribution, listed in alphabetical order.}, Bingjin Chen\footnotemark[1], Huang He\footnotemark[1], Xin Tian\footnotemark[1], Han Zhou\footnotemark[1],\\ Fan Wang, Hua Wu, Haifeng Wang, Wenquan Wu, Yingzhan Lin\\}
\affiliations{Baidu Inc., China\\
\{baosiqi, chenbingjin, hehuang, tianxin06, zhouhan05, \\
wang.fan, wu\_hua, wanghaifeng, wuwenquan01, linyingzhan01\}@baidu.com
}

\begin{document}

\maketitle

\begin{abstract}
In this work, we explore the application of PLATO-2 on various dialogue systems, including open-domain conversation, knowledge grounded dialogue, and task-oriented conversation. PLATO-2 is initially designed as an open-domain chatbot, trained via two-stage curriculum learning. In the first stage, a coarse-grained response generation model is learned to fit the simplified one-to-one mapping relationship. This model is applied to the task-oriented conversation, given that the semantic mappings tend to be deterministic in task completion. In the second stage, another fine-grained generation model and an evaluation model are further learned for diverse response generation and coherence estimation, respectively. With superior capability on capturing one-to-many mapping, such models are suitable for the open-domain conversation and knowledge grounded dialogue. For the comprehensive evaluation of PLATO-2, we have participated in multiple tasks of DSTC9, including interactive evaluation of open-domain conversation (Track3-task2), static evaluation of knowledge grounded dialogue (Track3-task1), and end-to-end task-oriented conversation (Track2-task1). PLATO-2 has obtained the 1st place in all three tasks, verifying its effectiveness as a unified framework for various dialogue systems. 
\end{abstract}

\section{Introduction}
The neural models in the conversational AI can be roughly divided into three categories: open-domain chatbot, knowledge grounded dialogue agent, and task-oriented dialogue system \cite{gao2018neural}. Due to the significant differences among these tasks, it is usually necessary to customize the modeling and training for each task. Recently, pre-trained language models have gained tremendous success in natural language processing \cite{devlin2019bert, brown2020language} and pioneering efforts have been made to pre-train dialogue generation models \cite{bao2019plato, zhang2019dialogpt}. However, there still lacks a unified pre-training framework which may effectively handle all these three conversational tasks. 

\begin{figure}
	\centering
	\includegraphics[width=0.48\textwidth]{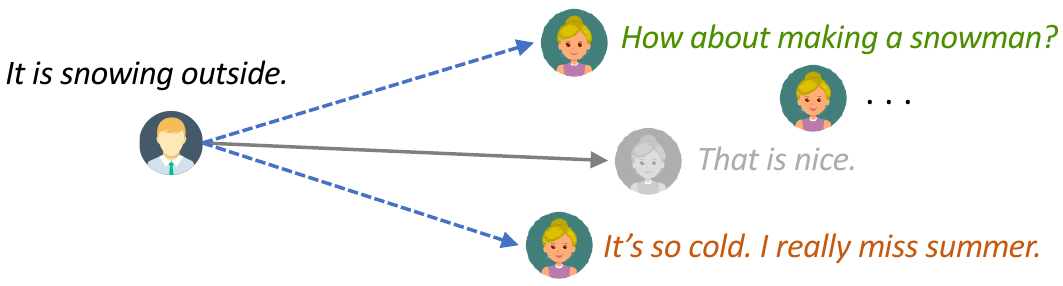}
	\caption{Toy example to show the one-to-one mapping (gray line) and one-to-many mapping (blue dashed lines) in open-domain conversations. Left: dialogue context; Right: candidate responses.}
	\label{fig:toy}
\end{figure}

\begin{figure*}
	\centering
	\includegraphics[width=\textwidth]{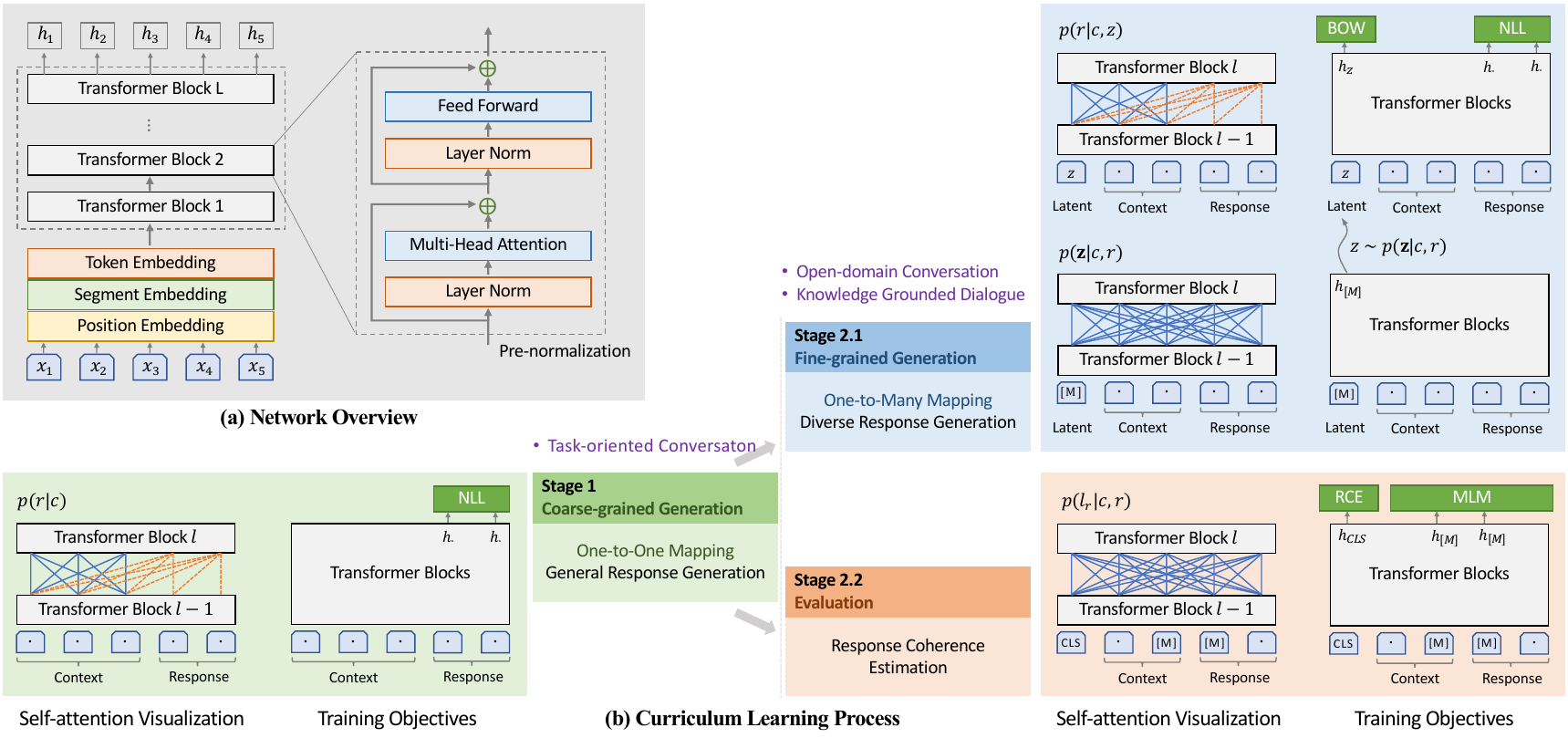}
	\setlength{\abovecaptionskip}{-4pt}
	\setlength{\belowcaptionskip}{-8pt}
	\caption{PLATO-2 illustration. (a) Network overview with the details of transformer blocks. (b) Curriculum learning process with self-attention visualization and training objectives.}
	\label{fig:network}
\end{figure*}

In this work, we will explore the application of PLATO-2 \cite{bao2020plato} on the aforementioned tasks, including open-domain conversation, knowledge grounded dialogue, and task-oriented conversation. PLATO-2 is initially designed as an open-domain chat-bot\footnotemark[1], trained via two-stage curriculum learning. In the first stage, a coarse-grained model is trained for general response generation under the simplified relationship of one-to-one mapping. In fact, one dialogue context might have multiple appropriate responses in open-domain conversations, as shown in the toy example of Figure \ref{fig:toy}. The one-to-one mapping network can only capture the common response patterns, resulting in general and dull responses during inference. As such, the curriculum learning continues to the next stage for high-quality response generation, as illustrated in Figure \ref{fig:network}. In the second stage, the discrete latent variable is encoded into the network for the one-to-many relationship modeling. Another fine-grained generation model and an evaluation model are further learned for diverse response generation and coherence estimation, respectively. The combination of fine-grained generation and evaluation helps PLATO-2 obtain new state-of-the-art results in open-domain conversations.
\footnotetext[1]{Source code and pre-trained models are released at \url{https://github.com/PaddlePaddle/Knover/tree/develop/projects/PLATO-2}.}

Similar to the open-domain conversation, the one-to-many mapping relationship also exists in knowledge grounded dialogue: given a dialogue context, multiple pieces of knowledge might be applicable for the response generation. Therefore, the one-to-many mapping models of the second stage can also be adapted for knowledge grounded dialogue. By expanding the network input with the knowledge segment, the background knowledge is encoded and grounded for response generation. Distinct from the open-domain conversation and knowledge grounded dialogue, there is a specific goal to be accomplished in task-oriented conversation. Accordingly, the conversation flow would become less diverse and concentrated on task completion. Therefore, the one-to-one mapping generation model of PLATO-2 is employed for the end-to-end task-oriented conversation. Given the dialogue context, this model will learn to generate the dialogue state, system action, and system response all together. 

To evaluate PLATO-2's performance on various dialogue systems, we have participated in multiple tasks of DSTC9 \cite{gunasekara2020overview}, including interactive evaluation of open-domain conversation (Track3-task2), static evaluation of knowledge grounded dialogue (Track3-task1), and end-to-end task-oriented conversation (Track2-task1). PLATO-2 obtains 1st place in all the three tasks, whose effectiveness and generalization are verified through these comprehensive evaluations. 

\section{PLATO-2}
The network overview and curriculum learning process of PLATO-2 are illustrated in Figure \ref{fig:network}. The network backbone is consisted of transformer blocks with pre-normalization \cite{vaswani2017attention, radford2019language}. And the network's input representation is the sum of token, segment and position embeddings. Distinct with conventional Seq2Seq approaches, PLATO-2 adopts the unified network \cite{dong2019unified, bao2019plato}, where transformer block parameters are shared across encoder and decoder. 

As shown in Figure \ref{fig:network}, there are two stages involved in the curriculum learning process. In the first stage, a coarse-grained generation model is learned under the simplified relationship of one-to-one mapping. Given one training sample of dialogue context and response $(c, r)$, the training objective is to minimize the negative log-likelihood (NLL) loss:
\begin{equation}\nonumber
\mathcal{L}_{NLL}^{\text{Baseline}}=-\mathbb{E}~\log p(r|c)
=-\mathbb{E}~ \sum_{t=1}^T~\log p(r_t|c,r_{<t})~,
\end{equation}
where $T$ is the length of the response and $r_{<t}$ denotes previous tokens. In order to obtain better language understanding, bi-directional self-attention is enabled within the context part, shown as blue lines. And for the auto-regressive generation, uni-directional self-attention is employed within the response part, shown as orange dashed lines. As discussed in the introduction, there exists the one-to-many relationship in open-domain conversations, i.e., one dialogue context may correspond to multiple appropriate responses. The one-to-one mapping network can only capture the typical patterns of diversified responses, resulting in general and dull responses during inference. Despite the problem of safe responses, the network is still highly effective in capturing the coarse-grained mapping relationship between dialogue context and response. 

To obtain high-quality responses for open-domain conversations, a fine-grained generation model and an evaluation model are further learned in stage 2. The discrete latent variable is encoded for the one-to-many relationship modeling, acting as a latent speech act. For the sake of accurate optimization, latent act recognition is first carried out to estimate the distribution of the latent variable w.r.t. the training sample $p(\mathbf{z}|c,r)$. The response is then generated with the sample latent variable $p(r|c,z)$, where $z\sim p(\mathbf{z}|c,r)$. The calculation of NLL loss becomes:
\begin{equation}
\begin{split}
\mathcal{L}_{NLL}^{\text{Generation}}&=-\mathbb{E}_{z\sim p(\mathbf{z}|c,r)} ~\log p(r|c,z)\\
&=-\mathbb{E}_{z\sim p(\mathbf{z}|c,r)} \sum_{t=1}^T~\log p(r_t|c,z,r_{<t})
\end{split}
\raisetag{2.6\baselineskip}
\end{equation}
where $z$ is one $K$-way categorical variable $z\in \{1,\cdots,K\}$. Given that the sampling operation is not differentiable, we approximate it with Gumbel-Softmax \cite{jang2016categorical}. Besides the classical NLL loss, the bag-of-words (BOW) loss \cite{zhao2017learning} is also employed to facilitate the training process of latent variable:
\begin{equation}
\begin{split}
\mathcal{L}_{BOW}^{\text{Generation}}&=-\mathbb{E}_{z\sim p(\mathbf{z}|c,r)}  \sum_{t=1}^T ~ \log p(r_t|c,z)\\
&=-\mathbb{E}_{z\sim p(\mathbf{z}|c,r)} \sum_{t=1}^T ~ \log \frac{e^{f_{r_t}}}{\sum_{v\in V} e^{f_v}}~,
\end{split}
\raisetag{2.6\baselineskip}
\end{equation}
where $V$ refers to the vocabulary, the function $f$ tries to predict all the words in the target response using the output embedding $h_z$. As compared with the NLL loss, BOW loss ignores the word order and forces the final latent embedding to capture the global information of the response. To sum up, the training objective of the fine-grained generation model is to minimize the integrated loss:
\begin{equation}
\mathcal{L}^{\text{Generation}}=\mathcal{L}_{NLL}^{\text{Generation}}+\mathcal{L}_{BOW}^{\text{Generation}}
\end{equation}

By assigning distinct values to the latent variable, the fine-grained generation model is able to produce multiple diverse responses. For selecting the most appropriate one from these candidate responses, the evaluation model is trained to estimate the coherence between each response and the given dialogue context. During training, the evaluation model needs to distinguish the golden response $r$ from the randomly selected negative response $r^-$. 
\begin{equation}\nonumber
\mathcal{L}_{RCE}^{\text{Evaluation}}=-\log p(l_r=1|c,r)
-\log p(l_{r^-}=0|c,r^-)
\end{equation}
Besides the response coherence estimation (RCE) loss, the conventional masked language model (MLM) loss \cite{devlin2019bert} is also included to retain the representation ability of the network. To sum up, the training objective of the evaluation model is to minimize the integrated loss:
\begin{equation}
\mathcal{L}^{\text{Evaluation}}=\mathcal{L}_{RCE}^{\text{Evaluation}}+\mathcal{L}_{MLM}^{\text{Evaluation}}
\end{equation}
During inference, conditioned on each latent value $z\in \{1,\cdots,K\}$, its corresponding candidate response is produced by the fine-grained generation model $p(r|c, z)$. The most coherent response can be selected in the following way:
\begin{equation}
    r^* = \max_{z \in \{1,\cdots,K\}}~ p(l_{r}=1|c,r)
\end{equation}
In addition to the above coherence estimation, two other approaches are commonly adopted for response selection: length-averaged log-likelihood and maximum mutual information. The length-averaged log-likelihood considers the forward probability $p(r|c)$, which tends to select those common and general responses. The maximum mutual information considers the backward probability $p(c|r)$, which favors those responses of high-overlap with the dialogue context. In comparison, the evaluation model $p(l_r=1|c,r)$ considers the bi-directional information flow between the dialogue context and response, achieving better performance at selecting coherent responses.

PLATO-2 learns gradually from coarse-grained general response generation to fine-grained diverse response generation via this curriculum learning process. Besides, the evaluation model further selects the most coherent response from multiple candidate responses. This combination of fine-grained generation and evaluation helps PLATO-2 obtain high-quality responses in open-domain conversations.

\section{Knowledge Grounded Dialogue}
\begin{figure}
	\centering
	\includegraphics[width=0.48\textwidth]{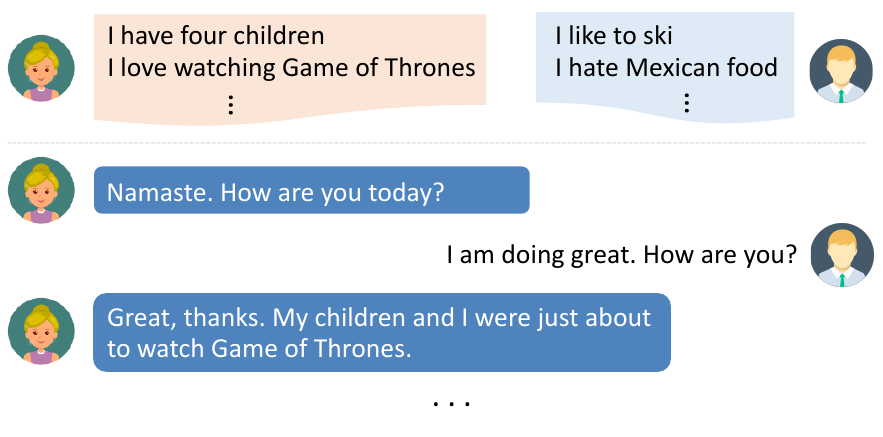}
	\caption{One example of knowledge grounded dialogue from Persona-Chat.}
	\label{fig:persona}
\end{figure}

Another common conversational task is knowledge grounded dialogue, where the response is generated based on the dialogue context and background knowledge. The background knowledge can come in a variety of forms, such as persona profiles \cite{zhang2018personalizing}, Wikipedia \cite{dinan2018wizard}, news articles \cite{gopalakrishnan2019topical}, and so on. One example from Persona-Chat is given in Figure \ref{fig:persona}. It can be observed that the response generation relies on not only the dialogue context but also the persona profiles. Similar to the open-domain conversation, there also exists the one-to-many mapping relationship in knowledge grounded dialogue \cite{kim2019sequential}: given a dialogue context, multiple pieces of knowledge might be applicable for the response generation. 

Within the PLATO-2 framework, the background knowledge can be encoded into the fine-grained generation and evaluation network straightforwardly by adding a segment of knowledge before the dialogue context. The learning of response generation becomes $p(r|k,c,z)$, where $k$ refers to the background knowledge. And the evaluation model $p(l_r=1|k,c,r)$ will consider the coherence with the dialogue context and the consistency with the background knowledge simultaneously. The fine-grained generation model produces diverse knowledge grounded responses and the evaluation model further selects out the most appropriate one from these candidates. 

\begin{table*}
	\centering
	\includegraphics[width=\textwidth]{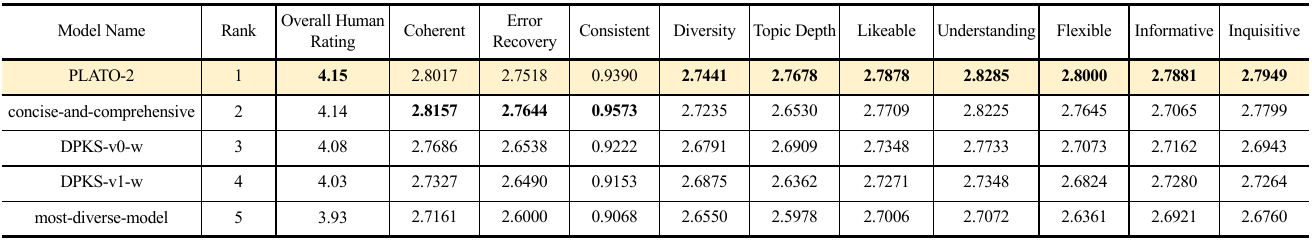}
	\caption{Human evaluation results on Track3-task2 interactive open-domain conversations, with the best value written in bold.}
	\label{tab:chat}
\end{table*} 
\begin{table*}
	\centering
	\includegraphics[width=\textwidth]{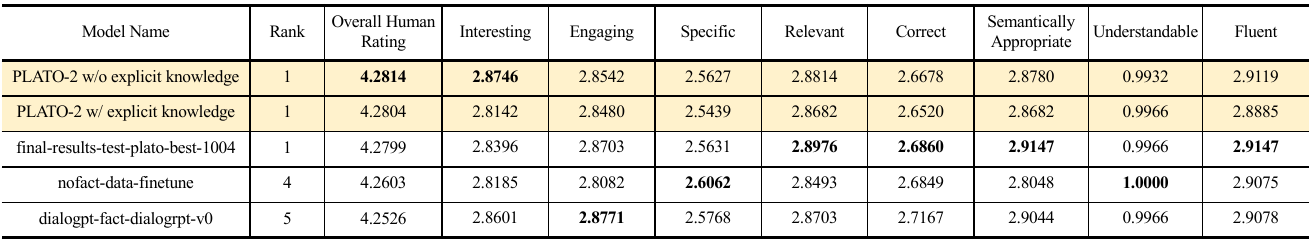}
	\caption{Human evaluation results on Track3-task1 static knowledge grounded dialogues, with the best value written in bold.}
	\label{tab:knowledge}
\end{table*} 
\section{End-to-end Task-oriented Conversation}
Conventional task-oriented dialogue systems usually adopt the pipeline architecture, including natural language understanding (NLU), dialogue state tracking (DST), dialogue policy, and natural language generation (NLG) modules. Recently, some works \cite{ham2020end, peng2020soloist} have been introduced for end-to-end task-oriented dialogue generation with pre-trained language models. In this section, we will discuss how to apply PLATO-2 on end-to-end task-oriented conversations. 

Distinct from open-domain conversation and knowledge grounded dialogue, the task-oriented conversation is supposed to accomplish a particular goal. Therefore, the semantic mapping between dialogue context and response would be less diverse. To this end, the one-to-one mapping generation model in stage 1 is employed for task-oriented conversation. Even with this powerful pre-trained generation model, it is still challenging to carry out end-to-end task-completion conversations. Firstly, the generation model needs to find out an effective way to interact with the external database. It is necessary to retrieve relevant information from the database for response generation, such as retrieving the candidates meeting the current user's criteria. Secondly, it is crucial for task completion to extract the entity precisely from the conversation. However, the entity name is non-categorical, and the user might mention it in various forms. Thirdly, the user's requests might be ambiguous, and the model has difficulties in capturing the user's real needs. Taking the utterance ``I want to find a hotel to stay" as an example, it is hard to tell whether the user wants to find a place to stay or the user wants to find a hotel instead of a guesthouse to stay. 

To tackle the above problems, several techniques are employed in our work. Firstly, the interaction with the external database is enabled through dialogue state estimation \cite{ham2020end} and a flexible two-phase generation process is adopted to produce the final response. In the first phase, the model generates the dialogue state, system action, and system response simultaneously. The dialogue state will be used as a constraint for database query, and the system action can be refreshed according to the queried results. If there is any update about the system action or no candidate found from the queried results, the second phase generation will be carried out to produce the final response. Secondly, to boost the extraction of entity names, we employ fuzzy matching between the dialogue context and database, where special tokens \texttt{<name/>} and \texttt{</name>} will be added around the candidate entity names. Through this enhanced presentation, our model achieves better accuracy and generalization in entity detection. Thirdly, to deal with the ambiguous requests, active clarification is introduced by raising one clarifying question towards the user, such as ``would you like a guesthouse or a hotel". With active clarification, the model can capture the user's real needs under ambiguous scenarios. The above three techniques -- effective interaction with an external database, improved entity representation, and active clarification, help PLATO-2 achieve a better success rate and user experience in task-oriented conversations. 

\begin{table*}
	\centering
	\includegraphics[width=\textwidth]{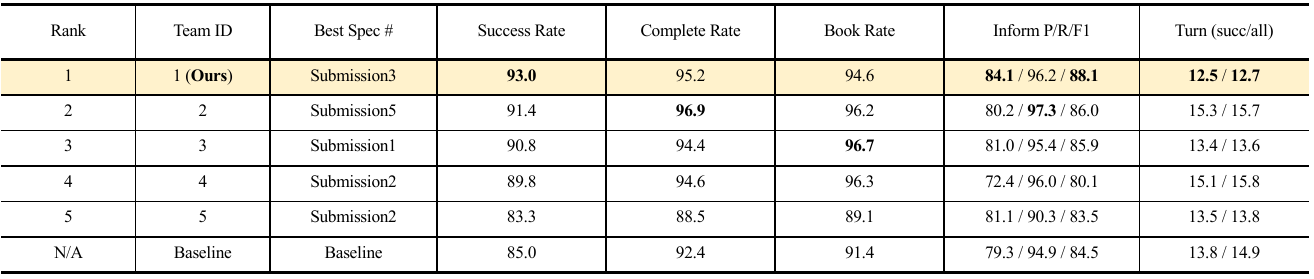}
	\caption{Automatic evaluation results on Track2-task1 end-to-end task-oriented conversations, with the best value written in bold.}
	\label{tab:task_auto}
\end{table*} 
\begin{table*}
	\centering
	\includegraphics[width=\textwidth]{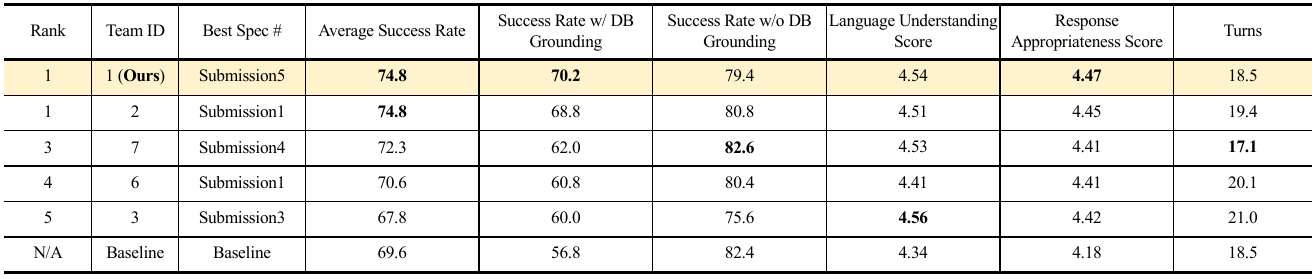}
	\caption{Human evaluation results on Track2-task1 end-to-end task-oriented conversations, with the best value written in bold.}
	\label{tab:task_human}
\end{table*} 
\section{Experiments}
For the comprehensive evaluation of PLATO-2, we have enrolled in multiple tasks of DSTC9 \cite{gunasekara2020overview}:
\begin{itemize}
    \item Track3-task2 interactive evaluation of open-domain conversation;
    \item Track3-task1 static evaluation of knowledge grounded dialogue;
    \item Track2-task1 end-to-end task-oriented conversation.
\end{itemize}
PLATO-2 is pre-trained with 684M (context, response) samples extracted from Reddit, and the vocabulary has 8k BPE subwords. All the models have 32 transformer blocks and 32 attention heads, with the hidden embedding dimension of 2048. For open-domain conversation and knowledge grounded dialogue, responses are generated with top-k sampling \cite{fan2018hierarchical}, where $k$ is set to 20. For task-oriented conversation, responses are produced with beam search, where the beam size is set to 5. Experimental details on each task will be discussed below.

\subsection{Open-domain Conversation}
Interactive open-domain conversation is the most challenging direction in dialogue systems. The users are free to talk about any topic and the system's replies are expected to meet a high standard on many aspects, including coherence, consistency, informativeness, engagingness, etc. Since PLATO-2 is initially designed as an open-domain chatbot, it can be applied directly in open-domain conversations. In DSTC9 Track3-task2, real internet users are attracted through Facebook advertising and communicate with the backend dialogue systems through DialPort \cite{zhao2016dialport}. The collected logs are then distributed to AMT workers for assessments. For each system, 200 interactive dialogues are collected for human evaluation. And for each dialogue, three crowd-sourcing workers are asked to annotate it from multiple aspects and provide an overall score. The human evaluation results are summarized in Table \ref{tab:chat}. PLATO-2 achieves the highest score of overall human rating and performs well on many evaluation metrics.

\subsection{Knowledge Grounded Dialogue}
In DSTC9 Track3-task1, experiments are carried out on Topical-Chat \cite{gopalakrishnan2019topical}, which is a large-scale dataset on knowledge grounded dialogue. For background knowledge, there are 300 entities in Topical-Chat, and each entity is associated with several short facts or articles. For each conversational turn, several relevant facts are provided, and the system can leverage these facts for response generation. As large-scale pre-trained models are capable of packing knowledge into the parameters \cite{roberts2020much}, we test two experimental settings: PLATO-2 with and without explicit knowledge. In the first setting, the given relevant facts are appended before the dialogue context, and the model learns the response generation based on explicit knowledge $p(r|k,c,z)$. In the second setting, the model tries to encode the knowledge into the network implicitly and learn the knowledge grounded response generation directly $p(r|c,z)$. 

In this task, systems need to produce the response given the dialogue context and relevant facts. During the evaluation, 100 randomly selected samples are distributed to AMT workers for assessments. For each conversational turn, three crowd-sourcing workers are asked to annotate it from multiple aspects and provide an overall score. The human evaluation results are summarized in Table \ref{tab:knowledge}. Three approaches are tied for the first place, where the top two are our submitted PLATO-2 without and with explicit knowledge. Given the name of the third approach, PLATO-2 might dominate the leaderboard in knowledge grounded dialogue.  
\begin{table*}
	\centering
	\includegraphics[width=\textwidth]{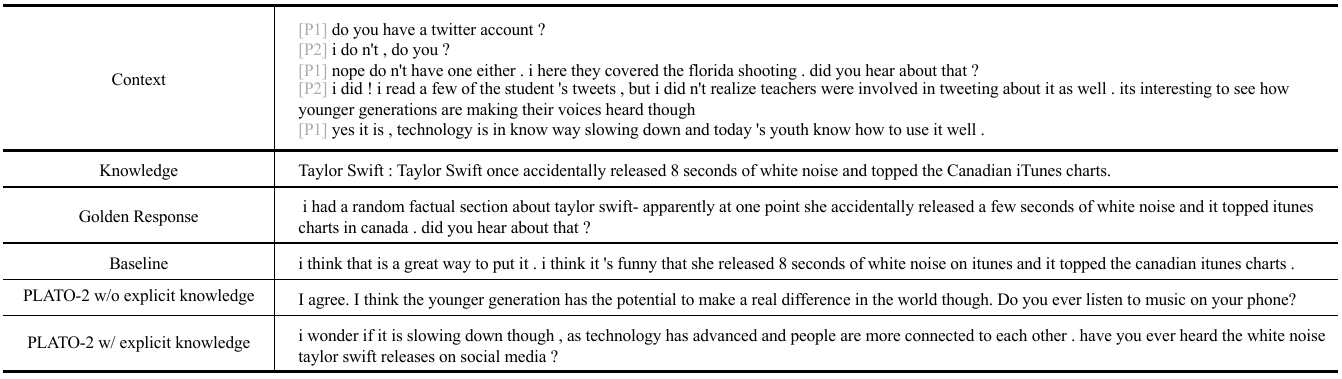}
	\caption{Case analysis on knowledge grounded dialogue.}
	\label{tab:case_knowledge}
\end{table*} 
\begin{figure}
	\centering
	\includegraphics[width=0.48\textwidth]{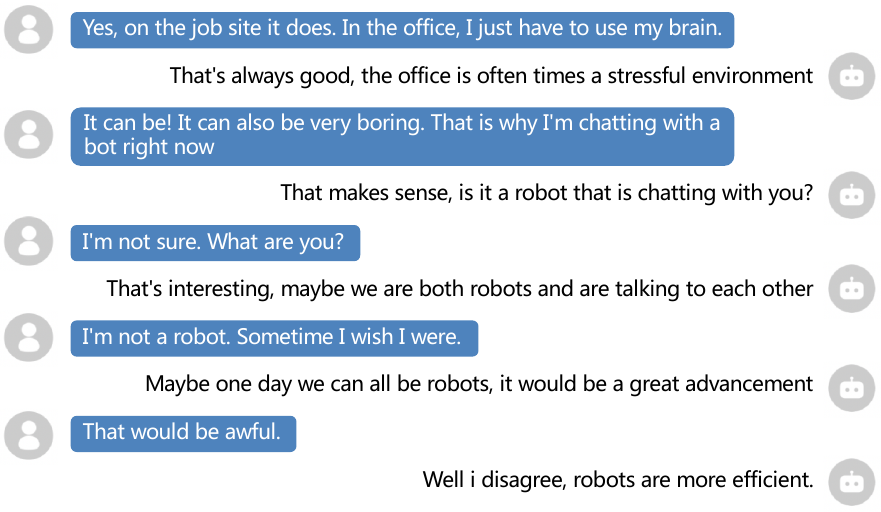}
	\caption{Case analysis on open-domain conversation.}
	\label{fig:case_chat}
\end{figure} 

\subsection{End-to-end Task-oriented Conversation}
In DSTC9 Track2-task1, the end-to-end task-oriented conversation is carried out on MultiWOZ 2.1 \cite{eric2020multiwoz}, which is a multi-domain dialogue dataset on travel planning. In the experiments, we employ the one-to-one mapping model of PLATO-2 and fine-tune it with the aforementioned three techniques. In this task, the ConvLab-2 platform \cite{zhu2020convlab} provides the user simulator for automatic evaluation. The evaluation results are summarized in Table \ref{tab:task_auto} and our approach ranks the 1st with the highest success rate. 

\begin{figure}
	\centering
	\includegraphics[width=0.48\textwidth]{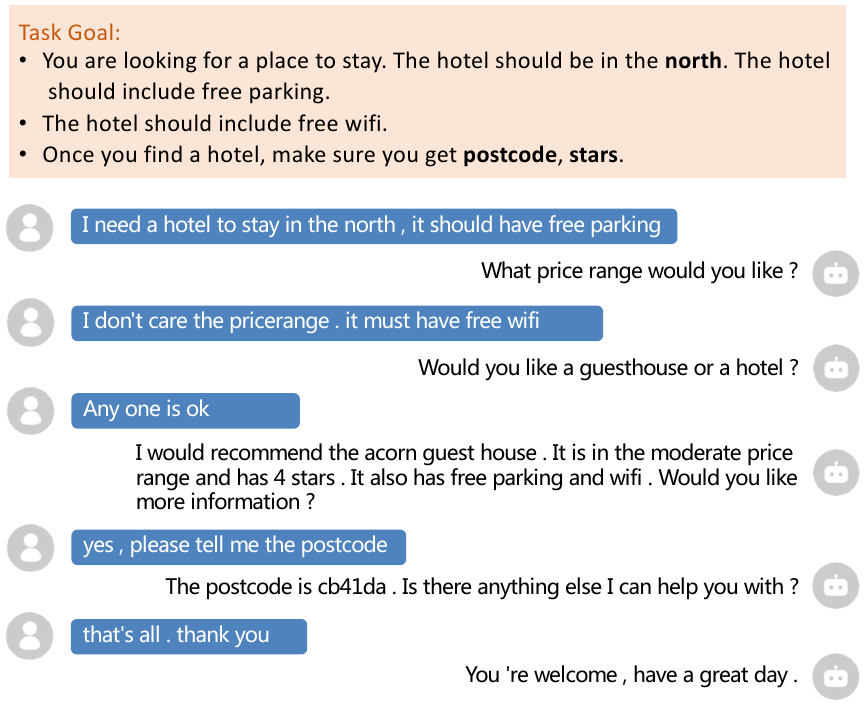}
	\caption{Case analysis on task-oriented conversation.}
	\label{fig:case_task}
\end{figure} 
Aside from the automatic evaluation, AMT workers are asked to communicate with the systems for task completion. When the conversation is finished, AMT workers need to give evaluation scores on several aspects. The human evaluation results are summarized in Table \ref{tab:task_human}. The average success rate is calculated as the average value of the following two metrics: the success rate without database grounding and the success rate with database grounding. The success rate without database grounding is based on the AMT worker's annotation during communication (success or fail). In fact, AMT workers do not know whether the provided values from the system are consistent with the database or not. In comparison, the success rate with database grounding is a more strict and practical metric. The dialogue is considered as a success if and only if: 1) AMT worker marks the dialogue as success; 2) the provided request slot values plus inform slot values from the system can be found in the database. Our approach achieves the highest score on success rate with database grounding. The first two approaches are placed as co-champion based on the average success rate in the final ranking. 

\subsection{Discussions}
To further dissect the performance of PLATO-2, several cases from various conversations are provided for analysis. As shown in Figure \ref{fig:case_chat}, one dialogue snippet is selected from the interaction between a real user and our system. In the DialPort platform, users are informed in advance that they will communicate with AI bots. This dialogue snippet demonstrates that PLATO-2 is able to produce coherent and engaging responses in open-domain conversation. For knowledge grounded dialogue, one example is selected to display in Table \ref{tab:case_knowledge}. As compared with the golden and baseline responses, the responses generated by PLATO-2 are more coherent with the dialogue context. Instead of changing the topic suddenly or copying the given facts directly, PLATO-2 absorbs the knowledge and conveys the information in a natural way. For task-oriented conversation, one dialogue snippet with the corresponding goal is selected and shown in Figure \ref{fig:case_task}. The user is asked to interact with the system to accomplish a specific goal. The system needs to find out the entity that satisfies the user's requirements. As exhibited in the case, the system actively communicates with the user to narrow down the scope of candidates and successfully returns the required information. 

Despite the effectiveness on multiple conversational tasks, PLATO-2 still suffers from several limitations of general dialogue models, including factual error, logic inconsistency, toxic and biased language, and so on. Recently, some pioneering works have been proposed to alleviate these problems. For example, the knowledge provenance from Wikipedia is provided in the retrieval-augmented generation \cite{lewis2020retrieval}. Some recipes are explored and discussed to increase the safety in open-domain chatbots \cite{xu2020recipes}. Future work will be carried out along these directions to boost the model's capacity. 

\section{Related Work}
Related works will be discussed on pre-trained dialogue generation models and task-oriented dialogue systems. 

Pre-trained language models have brought significant breakthroughs in natural language processing \cite{devlin2019bert, radford2019language, brown2020language}. To boost the performance of dialogue generation, DialoGPT \cite{zhang2019dialogpt} is trained on the basis of GPT-2 \cite{radford2019language} using Reddit comments. To obtain a human-like chatbot, Meena \cite{adiwardana2020towards} utilizes more social media conversations and scales up the network to 2.6B parameters. To strengthen the desirable conversational skills, Blender \cite{roller2020recipes} further fine-tunes the pre-trained model with human-annotated conversations. To tackle the one-to-many mapping problem, PLATO-2 \cite{bao2020plato} encodes discrete latent variable into the network and achieves new state-of-the-art results in open-domain conversations. In this work, we demonstrate that the one-to-many mapping models of PLATO-2 can be applied effectively on both open-domain conversation and knowledge grounded dialogue.

For task-oriented dialogue systems, conventional approaches \cite{young2013pomdp, henderson2014second, wen2015semantically} usually adopt pipeline modules, including natural language understanding (NLU), dialogue state tracking (DST), dialogue policy, and natural language generation (NLG). Recently, some end-to-end neural models \cite{wen2017network, ham2020end, peng2020soloist} have been introduced for task-oriented dialogue systems. The end-to-end system \cite{ham2020end} remains the core concepts of pipeline and generates the dialogue state, system action, and system response simultaneously. In this work, we demonstrate that the one-to-one mapping model of PLATO-2 can be adopted as a powerful basis. With enhanced entity representation and active clarification, PLATO-2 achieves new state-of-the-art results in task-oriented conversation.

\section{Conclusion}
In this work, we explore the application of PLATO-2 on various dialogue systems, including open-domain chit-chat, knowledge grounded dialogue, and task-oriented conversation. The training of PLATO-2 is carried out via two-stage curriculum learning. In the first stage, the network tries to fit the simplified one-to-one mapping between the dialogue context and response. In the second stage, the discrete latent variable is encoded into the network for the one-to-many mapping modeling. One fine-grained generation and one evaluation model are further learned for diverse response generation and coherence estimation. The model in the first stage is applicable to task-oriented conversation, while those models in the second stage are suitable for open-domain conversation and knowledge grounded dialogue. Comprehensive evaluations in DSTC9 demonstrate that PLATO-2 is an effective unified pre-training framework for conversational AI.

\section*{Acknowledgments}
We would like to thank the reviewers for their constructive suggestions; Jingzhou He, and Tingting Li for the help on resource coordination; Gaopeng Yong, Liankai Huang, and Hua Lu for their generous help. This work was supported by the Natural Key Research and Development Project of China (No. 2018AAA0101900).

\bibliography{bibtex}
\end{document}